\documentclass[runningheads]{llncs}
\usepackage{graphicx}
\usepackage{amsmath,amssymb} 
\usepackage{color}
\usepackage{subfig}
\usepackage{booktabs}
\usepackage{multirow}
\usepackage{url}
\usepackage{etoolbox}
\usepackage{float}
\usepackage[width=122mm,left=12mm,paperwidth=146mm,height=193mm,top=12mm,paperheight=217mm]{geometry}
\usepackage[font=small]{caption}

\def\cut#1{{}}
\DeclareRobustCommand\onedot{\futurelet\@let@token\@onedot}
\def\@onedot{\ifx\@let@token.\else.\null\fi\xspace}
\def\etal{\emph{et al}.}
\def\tabspacing {2pt}

\begin{document}
\pagestyle{headings}
\mainmatter

\title{A Simple Hierarchical Pooling Data Structure for Loop Closure} 

\titlerunning{A Simple Hierarchical Pooling Data Structure for Loop Closure}

\authorrunning{Fei \etal}

\author{Xiaohan Fei,
Konstantine Tsotsos,
Stefano Soatto}



\institute{UCLA Vision Lab\\
        University of California, Los Angeles\\
        \email{ \{feixh, ktsotsos, soatto\}@cs.ucla.edu}
}

\maketitle

\begin{abstract}
We propose a data structure obtained by hierarchically pooling Bag-of-Words (BoW) descriptors during a sequence of views that achieves average speedups in large-scale loop closure applications ranging from $2$ to $20$ times on benchmark datasets. Although simple, the method works as well as sophisticated agglomerative schemes at a fraction of the cost with minimal loss of performance. 

\end{abstract}
\section{Introduction}
We tackle the problem of {\em loop closure} in vision-based navigation. This is a particular classification task whereby a training set of images is indexed by location, and given a test image one wants to query the database to decide whether the former is present in the latter, and if so return the indexed location. This is closely related to {\em scene recognition}, where the focus is on a particular instance, as opposed to an object class (we want to determine whether we are at particular intersection in a given city, not whether we are at {\em some} intersection of {\em some} urban area). As such, test images are only subject to {\em nuisance variability} due to viewpoint, illumination and partial occlusion from moving objects, but otherwise there is no {\em intrinsic} (intra-class) variability.

The state-of-the-art for image retrieval is based on convolutional neural network (CNN) architectures, trained to marginalize nuisance and intrinsic variability. \cut{While some of this variability is managed by design, for instance through the use of linear convolutions and pooling, the rest is {\em learned away} by means of large annotated datasets. Unfortunately, existing convolutional architectures are not well matched to the loop closure problem: }In a discriminatively trained network, the compositionality property afforded by linear convolutions, while critical to model intra-class variability, is unhelpful for loop closure, as there is no intrinsic variability. At the same time, a CNN does not respect the topology of data space at higher levels of the hierarchy, since filters at any given layer are supported on the entire feature map of the previous layer. In loop closure, locality is key, and while one could retrieve from the feature map the locations that correspond to active units, this requires some effort~\cite{simonyan2014deep}. \cut{In this sense, CNNs are too much and too little for loop closure as they are not designed to specifically tackle the problem. It is not a coincidence that the state-of-the-art in loop closure consists of variants of BoW and inverted indices~\cite{galvez2011real,mur2015orb} that were ubiquitous in image retrieval before the advent of CNNs. which are powerful for object identification, not for applications like SLAM which we focus on in this paper. Unlike generic retrieval, however, BoW has remained the most popular.}

Given the critical importance of loop closure in location services ranging from smartphones to autonomous vehicles, we focus on its peculiarities, and attempt to harvest some of the components of neural networks to improve the state-of-the-art. Stripped of the linear convolutions (we do not need to model intrinsic variability) and ReLu\cut{(occlusions can be modeled as a subset selection, rather than a marginalization as suggested in~\cite{patel2015probabilistic})}, what we have left is a {\em hierarchical spatially pooled data structure built upon local photometric descriptors~\cite{girshick2014deformable,mahendran2015understanding}.} There are no filters, and no learning other than the trivial pooling of local descriptors. Motivated by this intuition, we propose a new hierarchical representation for loop closure, detailed in Sec.~\ref{sect-derivation}.


Loop closure is also closely related to location, or ``place,'' recognition~\cite{ulrich2000appearance,torralba2003context,cummins2009highly} and large-scale visual search~\cite{nister2006scalable,chum2007total,jegou2008hamming}, but with some important restrictions. 

First, both previous data (training images) and current (test, or query) data are usually available as time-indexed sequences, even if they are captured by different agents, and training images may be aggregated into a ``map''~\cite{jonesS09} or reduced to a collection of ``keyframes''~\cite{newcombe2010live}. Second, as a binary classification task (at each instant of time, a loop closure is either detected or not), the cost of  missed detections  and false alarms  are highly asymmetric: We pay a high price for declaring a loop closure that isn't, but there is minor harm in missing one, as temporal continuity affords many second chances in subsequent images. This is unlike large-scale image retrieval, where we wish to find what we are looking for (few missed detections, or high recall) even if we have to wade through some irrelevant hits (many false alarms, or low precision). 

Like image retrieval, however, the challenge with loop closure is scaling. In navigation applications, it may be hours before we return to a previously seen portion of the scene. Therefore, we have to store, and search through, hundreds of thousands to millions of images. Our goal in this paper is to {\em design a hierarchical data structure that helps speed up matching by leveraging on the two domain-specific constraints above:} temporal adjacency, and high precision. 

Assuming continuous trajectories, the first translates to proximity in pose space $SE(3)$ (position and orientation). For the second, the best trade-off with missed detections can be achieved by testing every datum in the training set via {\em linear search accelerated via an inverted index}. Our goal is to achieve similar performance at a fraction of the cost compared to inverted index search. This cannot be achieved in a worst-case setting.\cut{Indeed, the method we propose performs worse than inverted index search in the worst-case. However, this  is irrelevant in the application domain, since one can prune slow queries and accept a missed detection without appreciable consequences.} What matters instead is {\em average performance} trading off precision with computational cost. We evaluate such average performance empirically on the {\em KITTI}~\cite{geiger2013vision}, {\em Oxford}~\cite{cummins2009highly} and {\em TUM RGB-D}~\cite{sturm12iros} datasets, as well as demonstrate extensions to general image retrieval on the {\em ukbench}~\cite{nister2006scalable} and {\em INRIA Holidays}~\cite{jegou2008hamming} datasets. To demonstrate scalability, we also evaluate our algorithm on augmented datasets with around 40K images.

We propose a simple data structure based on hierarchical pooling of location likelihoods -- in the form of sample distributions of BoW descriptors -- with respect to the topology of pose space. In practice, this means simply constructing BoW descriptors, that represent the likelihood of the locations that generated them, and pooling them temporally in a fine-to-coarse fashion, either by averaging, summing, or taking the index-wise maximum. 

While averaging likelihoods may seem counter-productive, in Sec.~\ref{sect-derivation} we show it makes sense in the context of the classical theories of sampling and anti-aliasing. In Sec.~\ref{sect-expm} we show that, despite its simplicity, it works as well as sophisticated agglomerative schemes at a fraction of the effort. \cut{While it may seem that something this simple that works must have been done before, we could not find trace of it in the literature, reviewed in Sec.~\ref{sect-related}, perhaps owing to the counter-intuitive nature of building hierarchies of pooled likelihoods.}
\subsection{Related work}
\label{sect-related}
Loop closure is a key component in robotic mapping (SLAM)~\cite{williams09}, autonomous driving, location services on hand-held devices, and for wearables such as virtual reality displays. Loop closure methods can be roughly divided into 3 categories: appearance-only, map-only and methods in between. Appearance-only methods~\cite{cummins2009highly} are essentially large-scale image retrieval algorithms, influenced by~\cite{nister2006scalable} and more in general the literature of BoW object recognition and categorization~\cite{sivic2003video}. \cut{Recent advances in matching and verification for large-scale image retrieval include~\cite{furon2013using,Fragoso_2013_ICCV} and~\cite{arandjelovic2014dislocation}. } Map-only methods~\cite{kleinM07} use the data (images, but most often range sensors) to infer the configuration of points in 3D space, and then seek to match subsets of these points, often using variants of ICP~\cite{chetverikov2002trimmed} as a building block. These methods do not scale beyond a few hundreds of thousands of points, or thousands of keyframes, and are often limited to what is referred to as ``short-term'' loop closure~\cite{kleinM07}, necessary for instances when complete loss of visual reference occurs while tracking. There are also a variety of map-to-image and image-to-map~\cite{sattler2011fast} methods that show great promise, but have yet to prove scalability to the point where the map spans tens if not hundreds of kilometers~\cite{cummins2009highly}.

For scalability, the most common choice is to combine quantized local descriptors into a BoW and then use an inverted index. FAB-MAP~\cite{cummins2009highly} extends the basic setup by learning a generative model of the visual words using a Chow-Liu tree to model the probability of co-occurrence of visual words. \cut{This allows calling a loop closure even when very few common features are observed while rejecting false matches which might be similar in appearance but generated by different locations.}FAB-MAP 2.0 scales further by exploiting sparsity to make the inverted index retrieval architecture more efficient.\cut{Angeli \etal~\cite{angeli2008fast} adopt Recursive Bayesian filtering  with a simple voting scheme based on term frequency-inverted document frequency (TF-IDF)~\cite{blei2003latent,aizawa2003information,sivic2003video}.} Starting from~\cite{galvez2011real}, SIFT or SURF descriptors were replaced by more efficient binary descriptors such as BRIEF~\cite{calonder2010brief} and ORB~\cite{rublee2011orb} to achieve comparable precision and recall to FAB-MAP 2.0 with an order of magnitude speed increase. Several recent mapping and localization systems adopt it as a module, including~\cite{lim2014real} and ORB-SLAM~\cite{mur2015orb}. \cut{These also further incorporate ideas from graph optimization~\cite{kummerle2011g,strasdat2011double}.} 

In addition to the specific loop closure literature, general ideas from spatial data structures and agglomerative clustering~\cite{tishbyPB00} are also relevant to this work, including k-d trees~\cite{samet1990design}, dual trees and decision trees~\cite{geman1996active}, as well as data structures used for retrieval such as pyramid matching~\cite{grauman2005pyramid} and its spatial version~\cite{lazebnik2006beyond}. In more general terms, this work also relates to visual navigation and mapping, structure-from-motion, and location recognition, including the use of global descriptors~\cite{torralba2003context}.

Our method can be considered appearance-only, but it is loosely informed by geometry, in the sense that the scene domain (pose space) provides the topology with respect to which we pool descriptors. Also closely related to our approach are \cite{turcot2009better,torii2011visual}, which present techniques for merging only pairs of BoWs; in \cite{chum2007total} queries are expanded by using retrieved and verified images, which is orthogonal to and can be viewed as a {\it query-end} version of our method.\cut{To the best of our knowledge, our work is the first to pool BoWs hierarchically and specifically target search time in loop closure.}
\section{Methodology} 
\label{sect-derivation}
Since our focus is on a spatial structure that facilitates accelerated loop closure queries, we integrate components from recent state-of-the-art methods within our data structure and adopt such methods as a baseline, against which we compare our method. Specifically, we adopt~\cite{mur2015orb} as a baseline, consisting of a BoW where each word is an element of a dictionary of descriptors obtained off-line by hierarchical k-means clustering, with each word weighted by its inverse document frequency. FAST detectors~\cite{rosten2006machine} and BRIEF descriptors~\cite{calonder2010brief} are employed, and TF-IDF~\cite{blei2003latent,aizawa2003information,sivic2003video} is used to weigh the BoW relative to the inverse document frequency. This standard pipeline, with different clustering procedures to generate the dictionary and different features, comprises most basic large-scale retrieval systems, including appearance-only loop closure. However, the number of false alarms in large-scale settings is crippling, so temporal consistency and geometric verification are typically used as correction mechanisms. 
\subsection{Hierarchical testing}
\subsubsection{Construction of hierarchy}
Our data structure can be interpreted as a hierarchical version of TF-IDF. To illustrate the method, we first assume that every frame is a ``keyframe'' and therefore we have a time-series of BoWs, obtained as described above, and organized into a linear structure or {\em un-oriented list}, as we wish to retrieve frames regardless of the direction of traversal. Each node is associated with a histogram, in the form of a BoW, representing the likelihood of a pose $g(t) \in SE(3)$ (position $T(t) \in {\mathbb R}^3$ and orientation $R(t) \in SO(3)$) given the data (the image at time $t$, $I(t)$): $h^t \doteq {\rm BoW}(t) \sim p(I(t) | g(t))$, where the equivalence is up to normalization, and the density function is approximated with a histogram with $N$ bins, equal to the size of the dictionary.

\begin{figure}[htb]
\centering
        \subfloat[{\small Construction of hierarchy}]{
    {\includegraphics[width=0.45\linewidth]{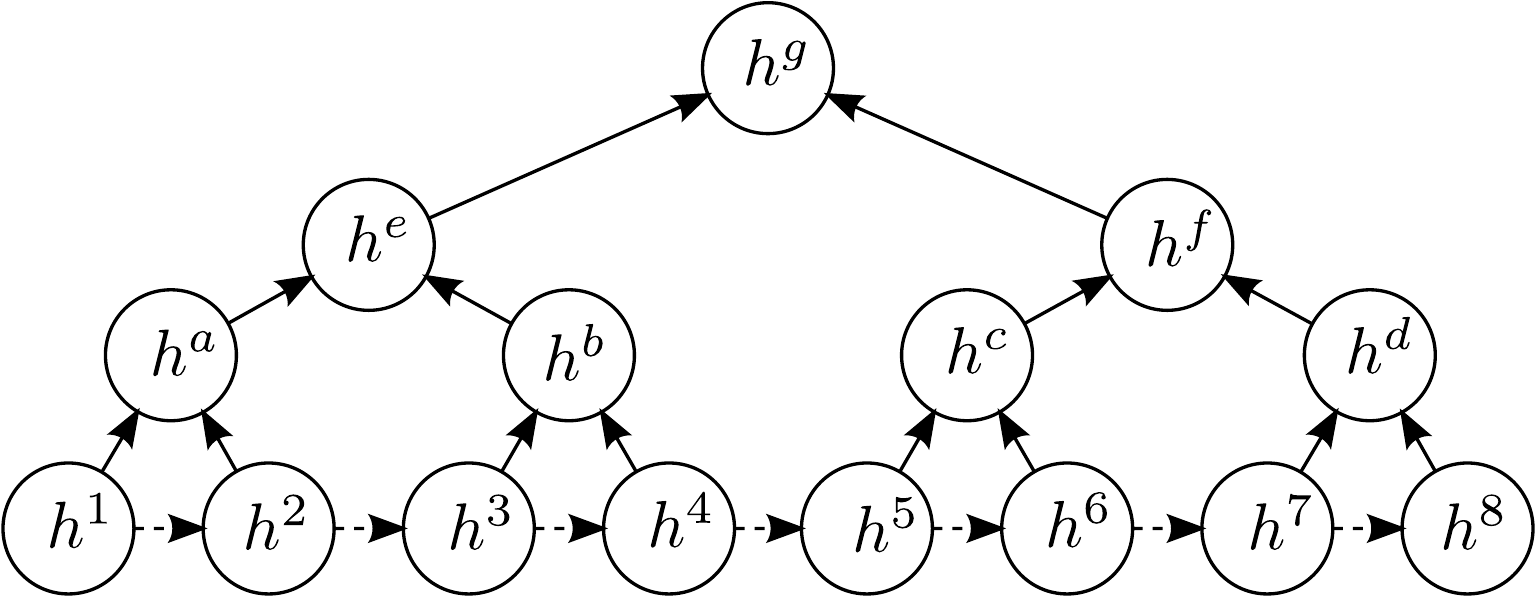}  }
    \label{fig-tree}}
    \hspace{10pt}
    \subfloat[{\small Hierarchical testing}]{
    {\includegraphics[width=0.45\linewidth]{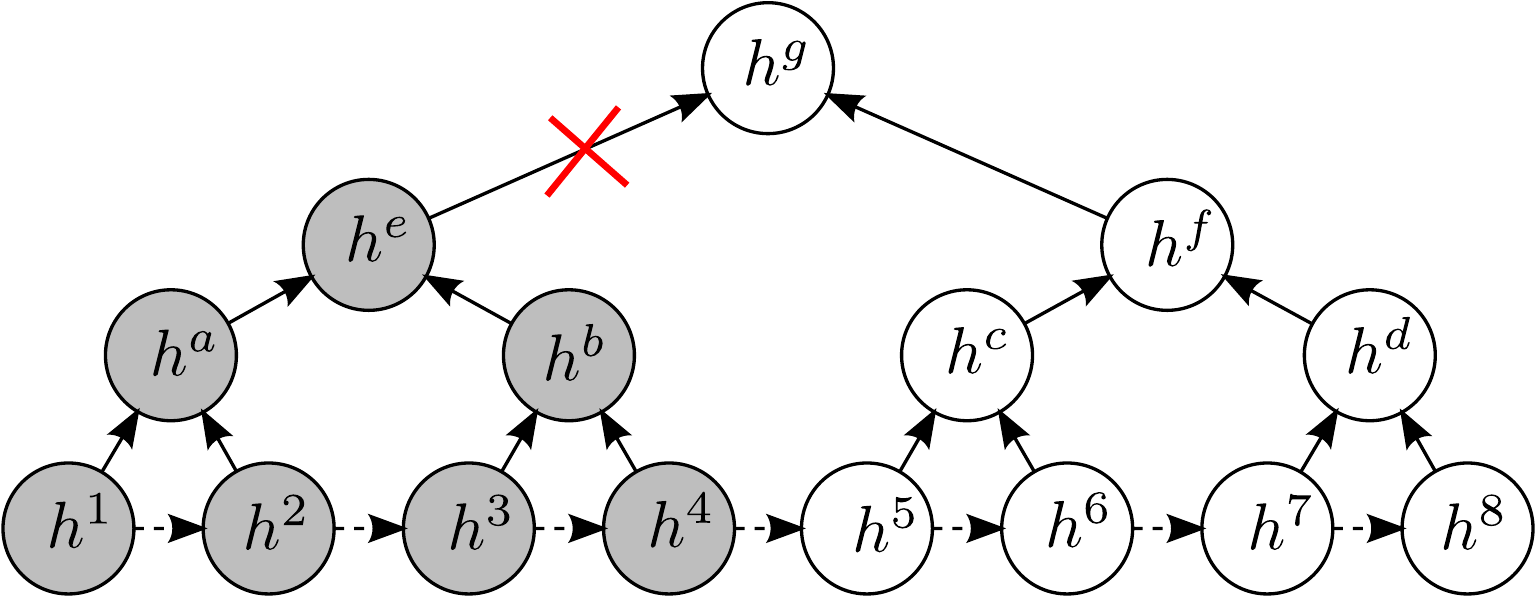}}
    \label{fig-testing} }
\caption{(a) {\bf Construction of hierarchy} for an 8-long sequence of (key)frames and constant branching factor of 2. Dashed lines indicate temporal order. (b) {\bf Hierarchical testing}: If $h^e$ does not score higher than the threshold, the whole sub-tree rooted at $h^e$ (shaded) will not be searched. In the case of sum- or max-pooling, this would {\it not} introduce loss of precision compared to searching only the lowest level nodes.}
\label{fig-illustration}
\end{figure}

We now construct a second level, or ``layer'', of the data structure, simply by pooling adjacent histograms (Fig.~\ref{fig-tree}). This is repeated for higher layers until either a maximum depth is reached, or until a single root node is left. Several standard choices for the pooling operation are available which allow us to trade off between precision and cost (Sec.~\ref{sect-expm}). Suppose $h^p$ is the parent histogram which has child histograms $\{ h^k \}, k=1,2\dots K $. Both $h^p$ and $h^k \in \mathbb{R}^N$. Mean- or average-pooling refers to $h^p=\frac{1}{K}\sum_{k=1}^{K}h^k$, sum-pooling refers to $h^p=\sum_{k=1}^{K}h^k$, and max-pooling refers to $h^p_i=\max_k\{ h^k_i \}$, where $i=1,2\dots N$. Once we have constructed the hierarchy for database histograms, raw histograms are used as queries for loop closure detection.

\subsubsection{Query processing}
\label{sect-testing}
 Similarities between pooled and query histograms are computed using the {\em intersection kernel}~\cite{swain1991color}, that is the area of the intersection of the two histograms. Thus, if $h^q$ (a query histogram) has bin values $h^q_1, \dots, h^q_N$, and similarly for $h^p$, we have that
\begin{equation}
{\mathbb I}(h^q, h^p) = \sum_{i=1}^N \min\{h^q_i, h^p_i\}
\end{equation}
The intersection kernel is related to many divergence functions~\cite{vasconcelos2004efficient} as well as to metrics used in optimal transport problems. 

Sum- and max-pooling operators have the following upper bound property when intersection kernel is applied:
For a query histogram $h^q$, a parent histogram $h^p$ and its child $h^k$ in the database,
 \begin{equation}
  \mathbb{I}(h^q, h^p)>\mathbb{I}(h^q, h^k)
 \end{equation}
therefore if $\mathbb{I}(h^q, h^p) < \tau$, $\mathbb{I}(h^q, h^k) < \tau$ must hold.

Since our goal is to search for the closest match, or at least for all matches that exceed a threshold $\tau > 0$ (we seek large values of $\mathbb I$), if ${\mathbb I}(h^q, h^p) < \tau$, the chance of any of $h^p$'s descendants exceeding the threshold is rare (or impossible, in the case of max- or sum-pooling as shown by the upper bound property), therefore we stop searching the sub-tree rooted at $h^p$ (Fig.~\ref{fig-testing}).

Therefore, search in a hierarchical TF-IDF setting simply boils down to {\em greedy breadth-first search, while maintaining an inverted index for each layer.} If only one layer is used, this reduces to standard linear search using an inverted index.

A key point is that with sum- or max-pooling, the proposed method {\em has exactly the same precision-recall behavior as standard inverted index search} while still achieving a substantial speedup. With mean-pooling, a large speedup can be achieved with only a minimal loss of precision (Sec.~\ref{sect-expm}).

Different trees with different depths and different branching factors can be constructed, trading off expected risk and computation time, characterized empirically in Sec.~\ref{sect-topology}. In addition to a fixed depth and branching factor, one could devise more clever schemes to determine the topology of the tree, discussed in Sec.~\ref{sect-fancy}. However, we find that the benefit is limited compared to the straightforward fixed-topology architecture.
\subsection{Keyframes and adaptive tree topology}
\label{sect-fancy}
So far we have assumed that the time-series of data $\{h^t\}_{t=1}^T$ is sampled regularly (at constant time or space intervals), but it can also be sampled adaptively, by exploiting statistics of the data stream to decide which samples, or {\em keyframes}, to use. The data structure above does not change, since all that is required is a topology or adjacency structure to construct the tree. 

Adaptive (sub)-sampling can be done in many ways, and there are a wide variety of standard heuristics for selecting keyframes. \cut{The most popular involve a combination of heuristics that includes the average number of tracked features and their overlap (new keyframes are spawned when the overlap of tracks with neighboring frames drop below a threshold), as well as hard gates on minimum and maximum time intervals, minimum and maximum distance in space or pose, between keyframes, including keyframes already existing in the map (but not temporally adjacent). While there is significant performance or computational cost modulation to be had,} Our goal here is not to determine the best method for selecting keyframes, but to focus on the data structure regardless of the sub-sampling mechanism. Consequently, we limit ourselves to constructing it either on the raw time series, or on any subsampling of it, as generated by standard keyframe selection methods.

Just like selecting keyframes, building the hierarchy can be understood as a form of (sub)-sampling. Regardless of whether subsampling is regular (as in building the tree above) or adaptive (as in selecting keyframes), classical sampling theory~\cite{smale2005shannon} suggests that what should be stored at the samples is {\em not} the value of the function, but the local average relative to the topology of the domain where the data are defined ({\em anti-aliasing}). This lends credence to the use of mean-pooling, which initially may seem counter-intuitive since our goal is to maintain high precision. 

In our case, the domain is time, or the order of keyframes, as a proxy of location in $SE(3)$. The range of the data is the space of likelihood functions, approximated by histograms $h^t$. Therefore, anti-aliasing simply reduces to averaging neighboring histograms. The study of the optimal averaging, both in terms of support and weights, is beyond our scope here, where for mean-pooling we simply average nearest neighbors in the tree topology relative to a uniform prior. We do not delve into considering more sophisticated anti-aliasing schemes, since we have found that simple topologies yield attractive precision-computational cost trade-off, which is unlikely to be significantly disrupted by fine-tuning the weights. 

The practice of averaging likelihood functions as a way of anti-aliasing descriptors has also been recently shown by~\cite{dongS15} in the context of pooling local descriptors for correspondences in wide-baseline matching. Our method can be considered an extension (or special case) where the correspondence and pooling are performed in time, and the descriptors are histograms of visual words, a mid-level representation, rather than histograms of gradient orientation, the result of low-level processing.

While the choice of heuristics for keyframe selection has no effect on our method, which can be applied to the raw time series or to the sequence of keyframes, the same (adaptive sampling) heuristics used to (down)-sample keyframes from the regularly sampled images could be used to aggregate nodes at one level into parents one level above. This would give rise to trees having different levels of connectivity at different layers, and indeed potentially at each node.

We have found that, in practice, these heuristics fail to yield significant performance improvements when compared to trees with fixed topology having constant splitting factors that match the average of their adaptive counter-part. Representative experiments are shown in Sec.~\ref{sect-topology}. 
\section{Evaluation} 
\label{sect-expm}
The most important evaluation for the proposed method is to test performance in-the-loop when incorporated into a real system (ORB-SLAM~\cite{mur2015orb}, in this case), discussed in Sec.~\ref{sect-baseline} where we find a $65\%$ reduction in mean query time with no loss in localization performance and no missed loop closures relative to the baseline. We investigate query-time reduction and precision-recall behavior while varying vocabulary size and tree topology in Sec.~\ref{sect-vocsize} and Sec.~\ref{sect-topology}, respectively. In Sec.~\ref{sect-synthex} we augment standard datasets to explore various test-time scenarios, and Sec.~\ref{sect-retrieval} presents a generalization of our method to other image retrieval tasks. Sec.~\ref{sect-datasets} discusses the datasets and methodology used throughout the evaluation.\cut{and Sec.~\ref{sect-assessment} summarizes the conclusions drawn from the experiments regarding construction of the hierarchical data structure.}

\subsection{Datasets and methodology for loop closure}
\label{sect-datasets}
We perform experiments using the common loop closure datasets of {\em KITTI}, {\em Oxford City Centre}, and {\em Oxford New College}~\cite{geiger2013vision,cummins2009highly}. The {\em KITTI} dataset consists of several sequences on the order of 1000 stereo pairs in length. To provide additional experimental evaluation at large scale, we augment {\em KITTI} by concatenating all sequences, to form the {\em concatenated KITTI} dataset consisting of approx. 40K images. For all sequences we construct the data structure using all frames unless otherwise noted, in which case we adopt the keyframe selection strategy of our baseline (Sec.~\ref{sect-baseline}). 

Unless otherwise stated, we build the hierarchical data structure using the left stereo images of the sequences (when stereo is available) and evaluate loop closure correctness using the provided ground truth poses. The evaluation protocol is as follows: traverse the sequence and insert BoW of images into the database incrementally, while using each image to query the database before it is added. Two images are regarded as a correct match if they were taken within 15 meters of each other. To avoid trivial matches, we prevent the query from matching temporally adjacent images. This evaluation protocol mimics loop closure in a practical SLAM system, which we test in Sec.~\ref{sect-baseline}.

To evaluate matching, missed detection and false alarms are traded off by an arbitrary choice of threshold, as in any detection algorithm. Since the threshold affects the average query time (we can make that quite short by choosing a threshold that yields no false alarms while rejecting every hypothesis) we must come to a reasonable choice. Unless otherwise stated, we adopt the following policy: We generate precision-recall curves on {\em KITTI} 00. Then, we select the smallest threshold that yields zero false alarms and use it on other sequences. Of course, that may yield a non-zero false alarm rate in datasets that are not used in setting the threshold, but this (as is customary) can be handled by verification steps afterwards. This is a limitation inherent to the choice of image representation, in this case Bag-of-Words, and not a sensitivity that our hierarchical data structure is designed to circumvent.
\subsection{In-the-loop with the baseline}
\label{sect-baseline}
We use components of ORB-SLAM~\cite{mur2015orb}, made available by the authors, as the baseline for our experiments. We use this as a \emph{black box} and implement our hierarchy atop its single-layer inverted index architecture for performing image queries. \cut{ORB-SLAM incorporates some recent techniques from binary feature descriptors and graph optimization. Modules in ORB-SLAM include FAST keypoint detection~\cite{tola2008fast}, ORB descriptors~\cite{rublee2011orb} and conversion to a BoW. In loop detection, to quickly retrieve BoW of previously visited places, an inverted index structure is maintained. ORB-SLAM also provides a heuristic to select keyframes, which we also use in our keyframe based experiments.} As a result, we also inherit some of the limitations of its components (e.g. keyframe selection, discriminability of quantized descriptors and BoW representations, sensitivity to matching threshold selection), which are common to the majority of SLAM systems.  

\begin{figure}[htb]
\centering
    \subfloat[{\small Scaling}]{
    {\includegraphics[width=0.41\linewidth]{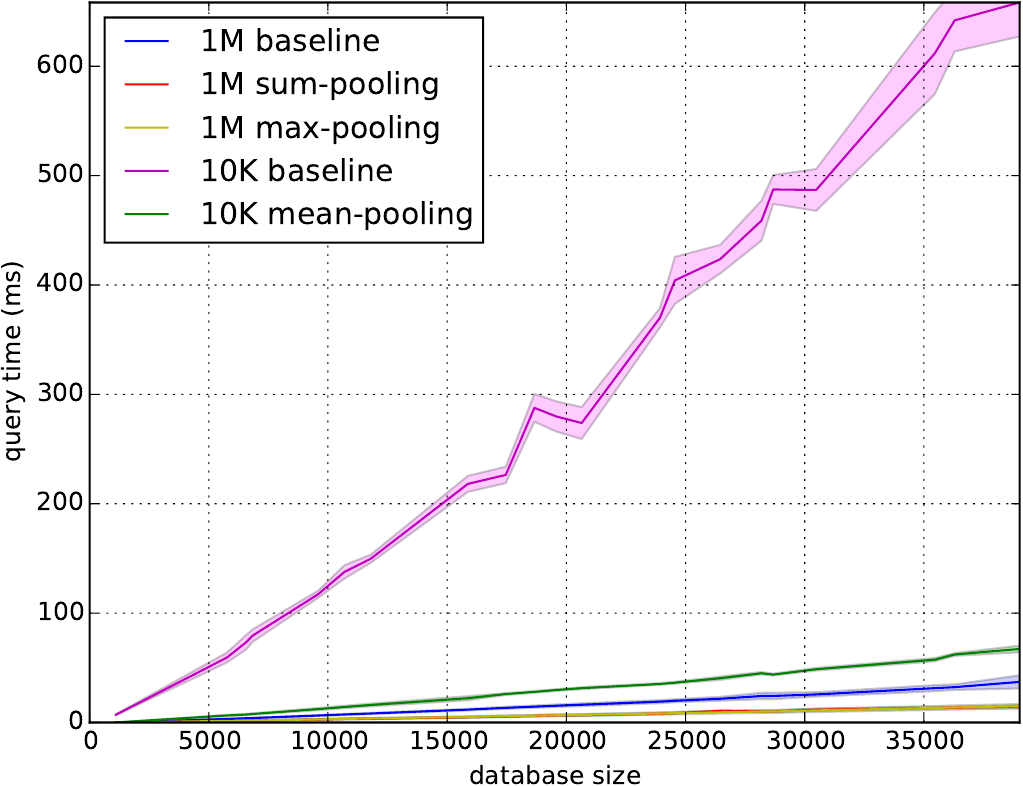} }
    \label{fig-kittiCat} }\hspace{20pt}
    \subfloat[{\small Comparison to ORB-SLAM}]{
    {\includegraphics[width=0.40\linewidth]{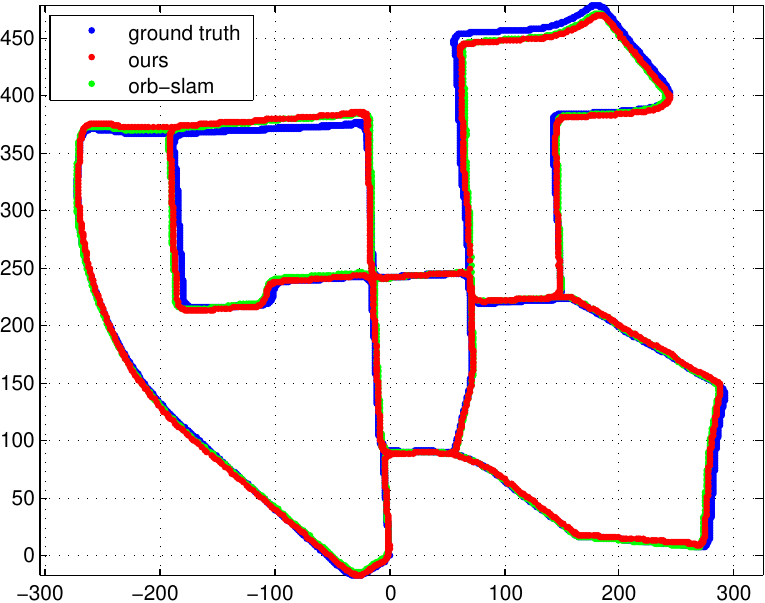} }
    \label{fig-orbslam} }
    \caption{(a) {\bf Scaling:} Timings for concatenated {\em KITTI} sequences (approx. 40K images) with 1M and 10K vocabularies. (b) Comparison to {\bf ORB-SLAM} with and without our data structure. Multiple trials yield nearly identical trajectories with and without our data structure, with no loop closures missed while achieving a 2-3x speedup.}
    \label{fig-misc}
\end{figure}

We first show that when using ORB-SLAM \emph{as is}, with no change in thresholds or tuning, a significant reduction in image query time can be achieved simply by applying our max-pooling hierarchy, which by construction achieves identical precision-recall performance to the original system, missing no loop closures that may be critical to pose-graph optimization algorithms. In Fig.~\ref{fig-orbslam}, we compare the trajectories estimated by ORB-SLAM with and without our max-pooling hierarchy on {\em KITTI}. Errors relative to ground truth are similar (within $1\sigma$ of each other over multiple trials); mean query times are reduced by 65\% (2.04ms from 5.80ms). No loop closures are missed by our max-pooling method that would not be missed without our data structure, confirming that improvement in speed comes at no loss of classification performance. In Fig.~\ref{fig-kittiCat} we show this speedup holds with increasing scale by showing query times for the {\em concatenated KITTI} dataset for different vocabulary sizes (Sec.~\ref{sect-vocsize}) and various pooling strategies using the methodology of Sec.~\ref{sect-synthex}. 

\cut{It should be noted that our policy to choose the threshold is common to BoW-based loop detection algorithms. In practice, one wants to find a reasonably large threshold such that the returned short list for each query is small and can be processed by verification procedures in a small amount of time (especially in applications such as real-time re-localization deployed on mobile devices).}
\subsection{Varying vocabulary size}
\label{sect-vocsize}
Some may argue that a speedup could be easily gained by just using a larger vocabulary. It is true that with a larger vocabulary, each visual word is associated with a much smaller list of documents in the inverted index system which leads to shorter query time. However, the vocabulary size should be determined by the performance of the specific task as well as the volume of the data and a larger vocabulary is not always better. A larger vocabulary has finer division of feature space compared to a smaller vocabulary but is also more sensitive to quantization errors (two slightly different images may have completely different histograms). In this case, mean-pooling may not be ideal as shown in Fig.~\ref{fig-prcurve00-1m} and~\ref{fig-prcurve02-1m}. However, sum/max-pooling can still be applied to gain further speedup while maintaining same precision-recall as shown in Fig.~\ref{fig-timings00-1m} and~\ref{fig-timings02-1m}, and also on augmented dataset as shown in Fig.~\ref{fig-kittiCat}.
\subsection{Varying tree topology}
\label{sect-topology}
\subsubsection{Variable depth and branching factor}
\label{sect-branching}
\begin{figure}[htb]
    \centering
    \subfloat[{\small {\em KITTI} 00-10K}]{
    {\includegraphics[width=0.43\linewidth]{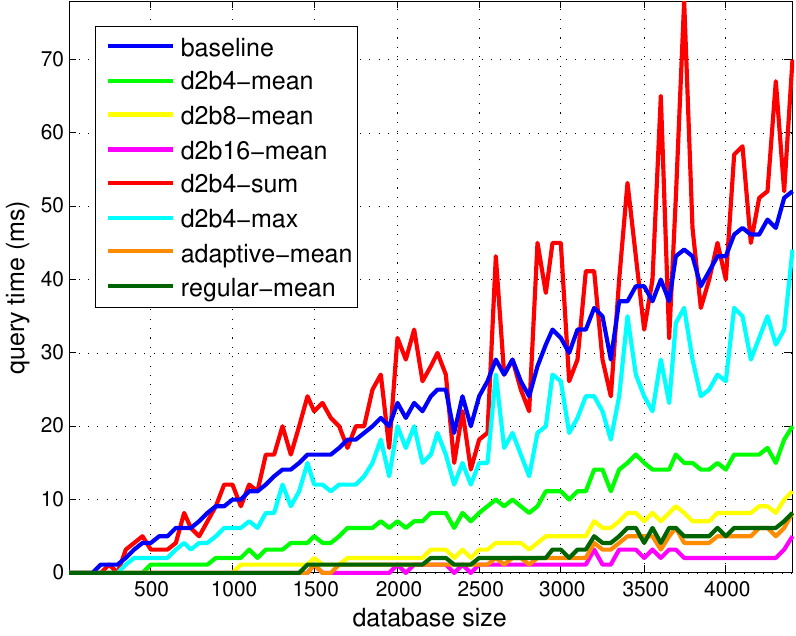} }
    \label{fig-timings00-10k} }
    \hspace{10pt}
    \subfloat[{\small {\em KITTI} 02-10K}]{
    {\includegraphics[width=0.44\linewidth]{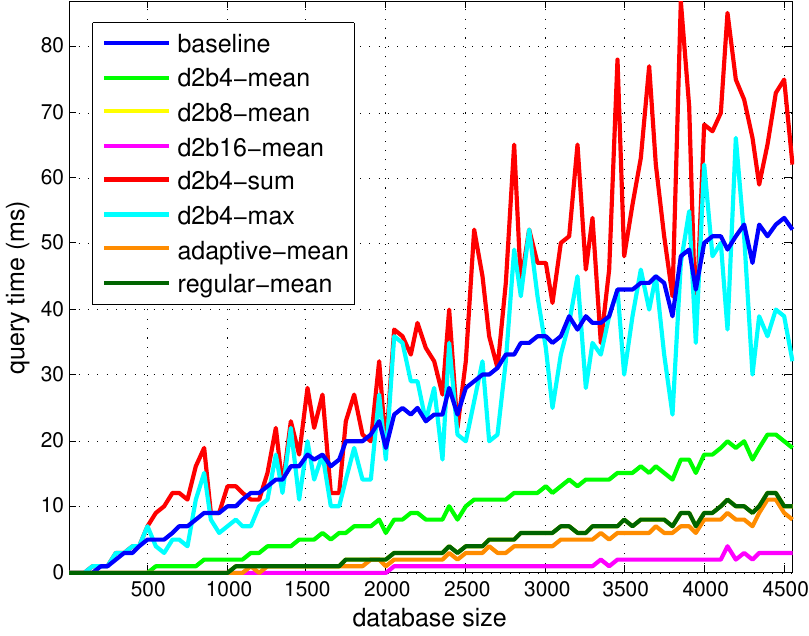} }
    \label{fig-timings02-10k} }\\
    \subfloat[{\small {\em KITTI} 00-1M}]{
    {\includegraphics[width=0.43\linewidth]{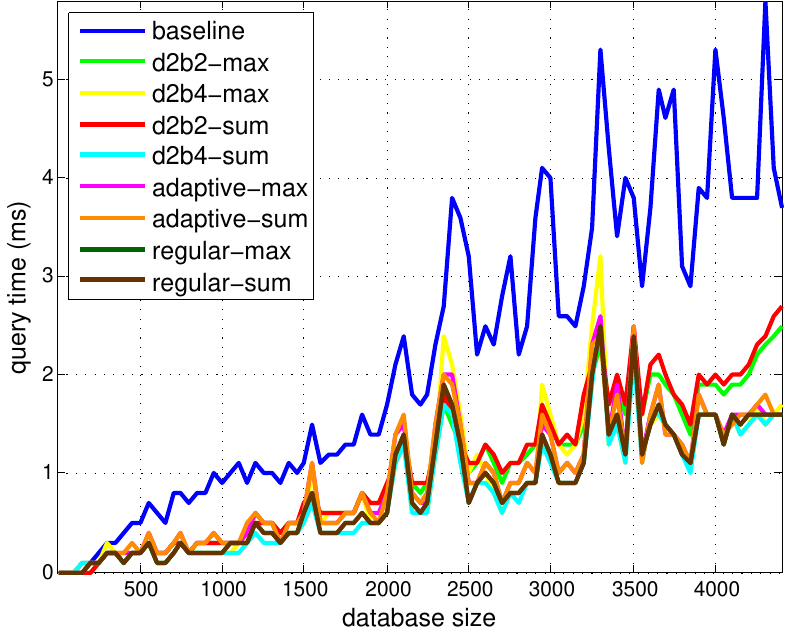} }
    \label{fig-timings00-1m} }
    \hspace{10pt}
    \subfloat[{\small {\em KITTI} 02-1M}]{
    {\includegraphics[width=0.45\linewidth]{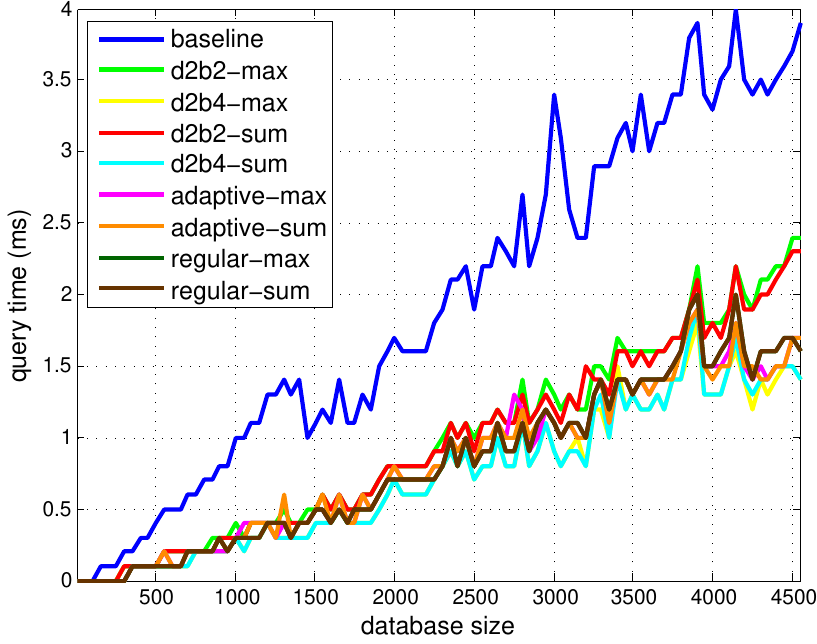} }
    \label{fig-timings02-1m} }
    \caption{Timings of baseline and proposed algorithm with different topologies and pooling strategies on {\em KITTI} dataset 00 and 02 using \emph{all} frames. $d_ib_j$~-X: a hierarchy with $i$ layers, a branching factor of $j$ and pooling strategy X. Adaptive sampling: spectral clustering in $SE(3)$. Regular sampling: sampling at the average rate of adaptive sampling scheme. Baseline: inverted index search. Two different vocabulary sizes (10K and 1M) are considered.}
    \label{fig-timings}
\end{figure}

\begin{figure}[htb]
    \centering
        \subfloat[{\small {\em KITTI }00-10K}]{
    {\includegraphics[width=0.4\linewidth]{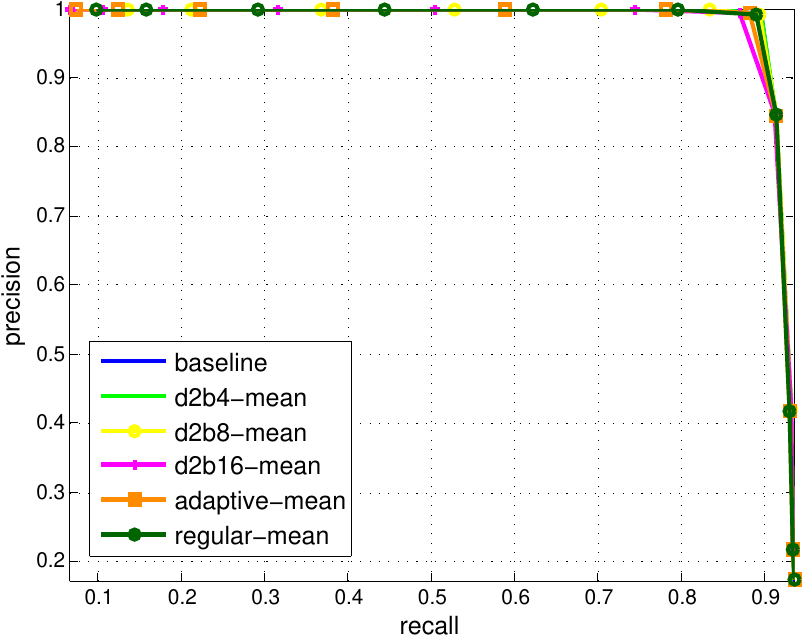}  }
    \label{fig-prcurve00-10k}}
    \hspace{15pt}
    \subfloat[{\small {\em KITTI }02-10K}]{
    {\includegraphics[width=0.4\linewidth]{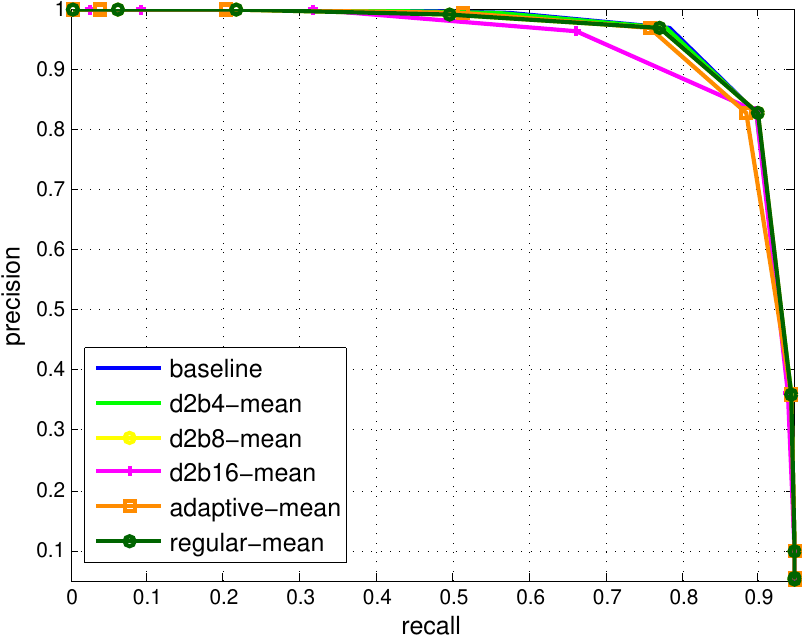}}
    \label{fig-prcurve02-10k} }\\
        \subfloat[{\small {\em KITTI }00-1M}]{
    {\includegraphics[width=0.4\linewidth]{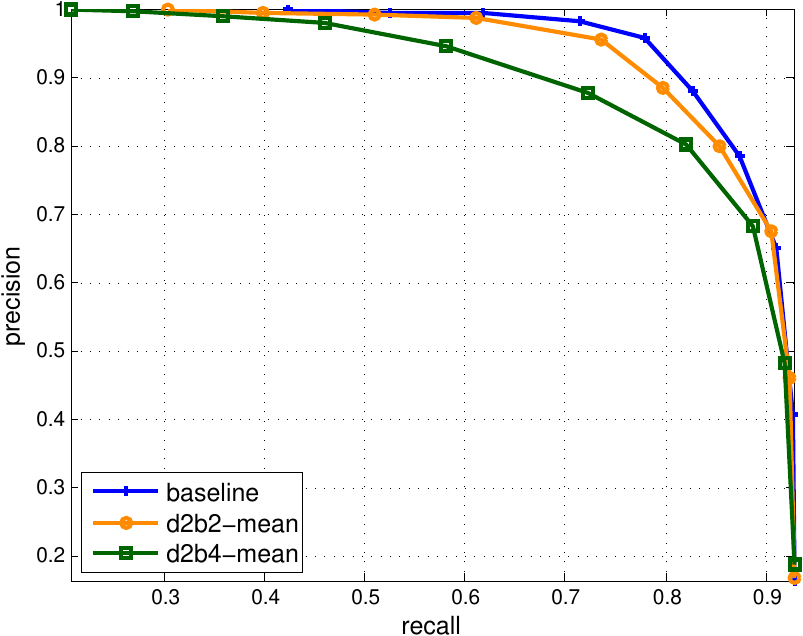}  }
    \label{fig-prcurve00-1m}}
    \hspace{15pt}
    \subfloat[{\small {\em KITTI }02-1M}]{
    {\includegraphics[width=0.4\linewidth]{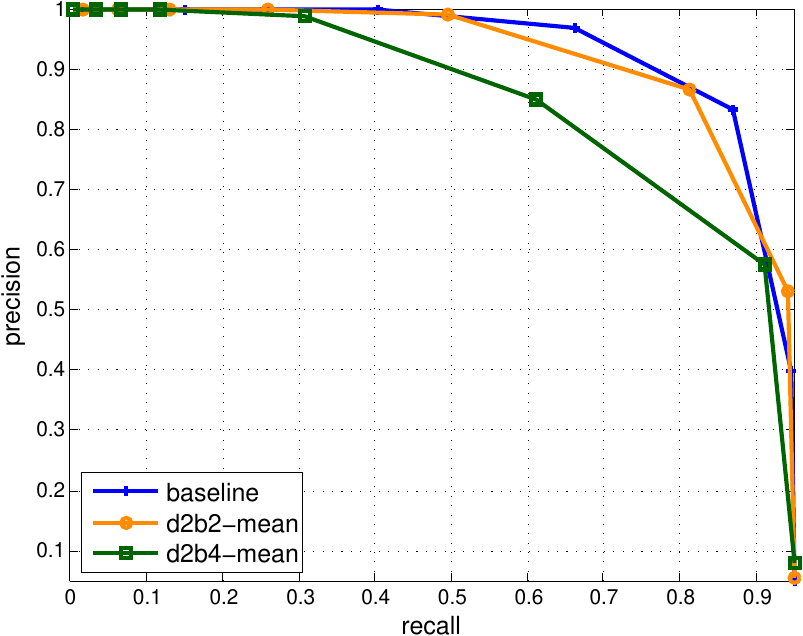}}
    \label{fig-prcurve02-1m} }
    \caption{Precision-recall curves of baseline and proposed algorithm with different topologies on {\em KITTI} dataset 00 and 02 using \emph{all} frames. Two different vocabulary sizes (10K and 1M) are considered. Notations have the same meanings as in Fig.~\ref{fig-timings}. }
\end{figure}

Fig.~\ref{fig-timings} shows timings of the baseline and our algorithm with different topologies and pooling schemes at the same threshold on two of the {\em KITTI} sequences with many loop closures. Only time to query the database is counted, time for feature extraction and descriptor quantization are excluded. Fig.~\ref{fig-prcurve00-10k} and Fig.~\ref{fig-prcurve02-10k} show precision-recall curves for the mean-pooling variants.
We use $d_ib_j$-X to denote a hierarchy with $i$ layers, a branching factor of $j$ and pooling strategy $X$. Note that for baseline and our proposed algorithm with configuration $d_2b_4$-mean and $d_2b_8$-mean, the precision-recall curves are nearly identical, while our approach is 2-5 times faster. For configuration $d_2b_{16}$-mean, while its performance is slightly worse, it achieves {\em an order of magnitude speedup} relative to the baseline.  

As mentioned in Sec.~\ref{sect-testing}, sum/max-pooling have {\it exactly the same precision-recall behavior} as the baseline. In these two datasets, sum/max-pooling are slightly slower than inverted index search. Since both of these operations rapidly reduce sparsity in the histograms, we expect slower performance relative to mean-pooling. However, sum/max-pooling have their advantages when a much larger vocabulary is used as shown in Sec.~\ref{sect-vocsize}.
\subsubsection{Adaptive domain-based clustering}
\label{sect-domain}
In addition to the baseline algorithm, we generate a second baseline by applying the same algorithm to keyframes, rather than to all stored images. In principle, the heuristics involved in the selection of keyframes could be propagated to all nodes of the data structure, as discussed in Sec.~\ref{sect-fancy}. However, our experiments indicate that this yields minor benefits compared to simple averaging. The second row of Tab.~\ref{tab-kittiKeyframe} shows average time-cost rate~\footnote{\scriptsize Time-cost rate is defined as the increase of query time per thousand (1k) images in the database. Average time-cost rate is the average of time-cost rates computed for all sequences in each dataset.} for searching via an inverted index among keyframes, which is worse than searching in a simple hierarchy built on raw images, as shown in the second row of Tab.~\ref{tab-kittiOverall}. A simple regular sampling strategy on top of keyframes can speedup searching by a large margin as shown in Tab.~\ref{tab-kittiKeyframe}. 

\begin{figure}[htb]
    \centering
    \subfloat
    {
      {\includegraphics[width=0.3\linewidth]{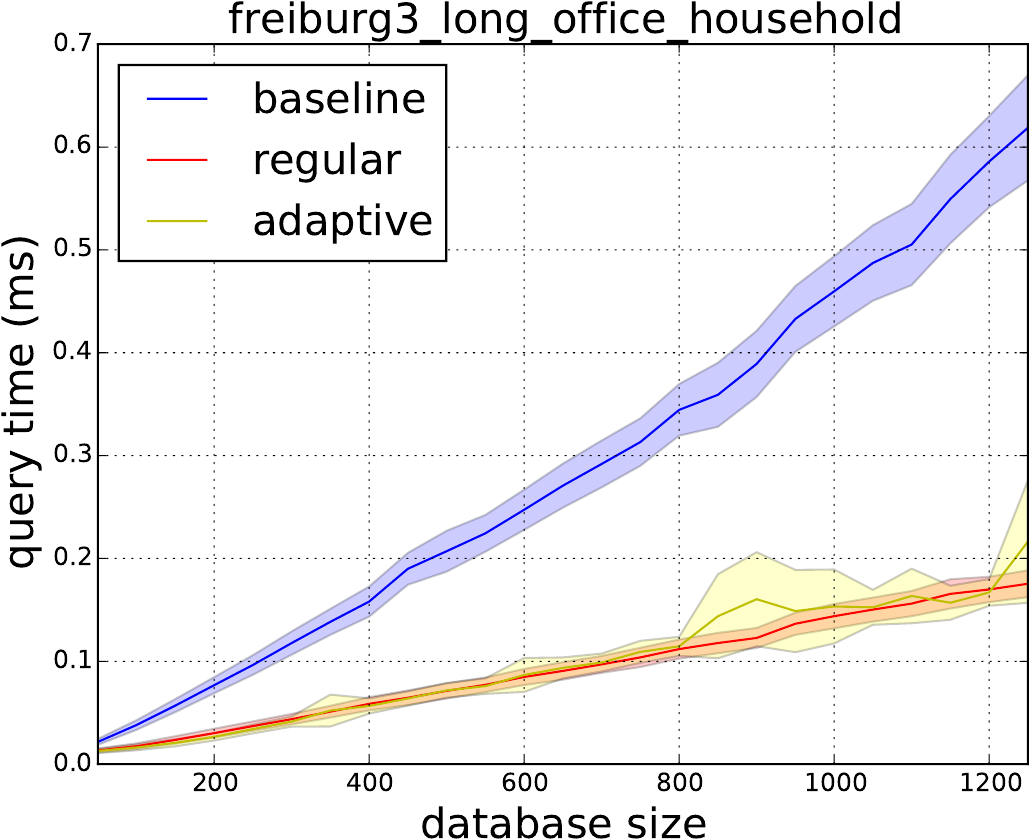} } 
    }
    \subfloat
    {
      {\includegraphics[width=0.3\linewidth]{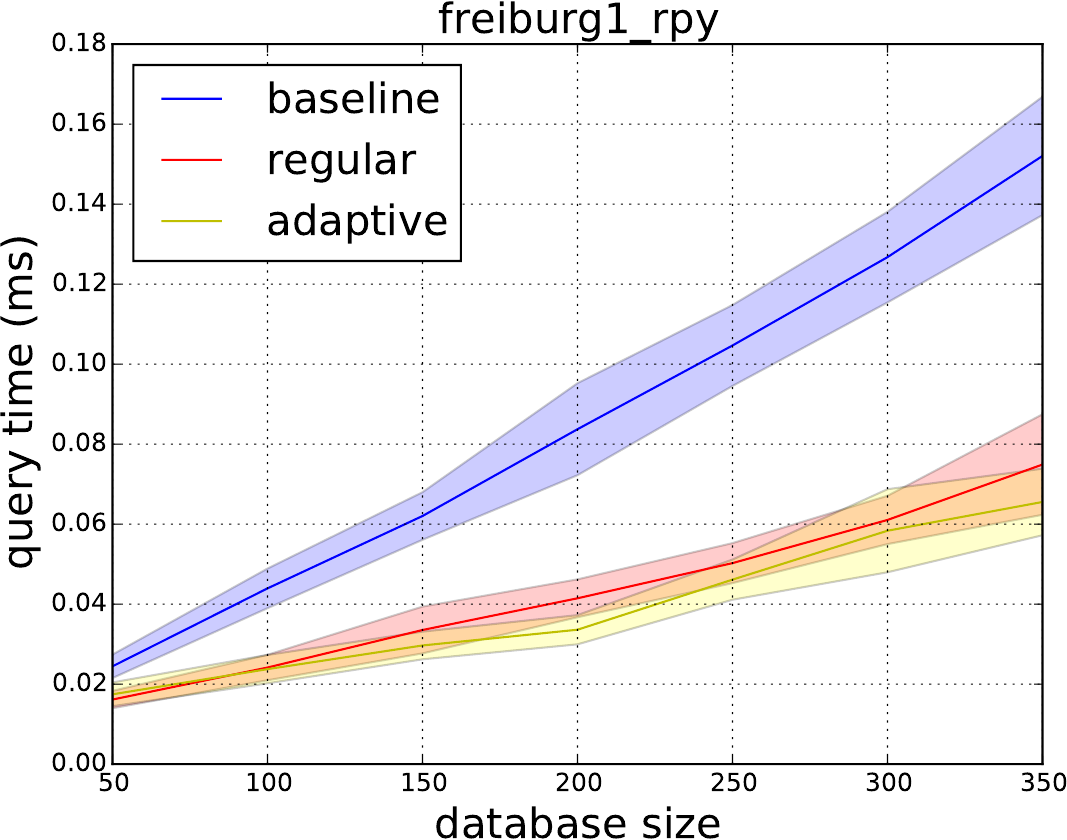} } 
    }
    \subfloat
    {
      {\includegraphics[width=0.3\linewidth]{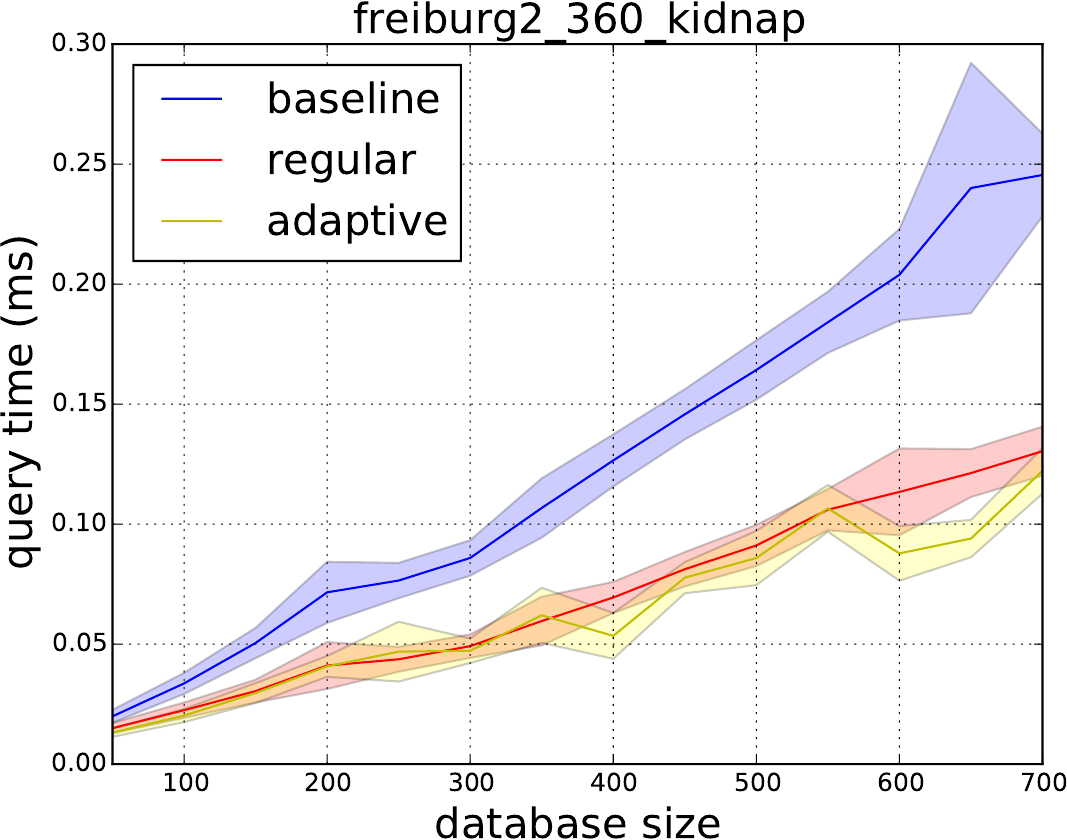} }
    }\\
    \subfloat
    {
      {\includegraphics[width=0.3\linewidth]{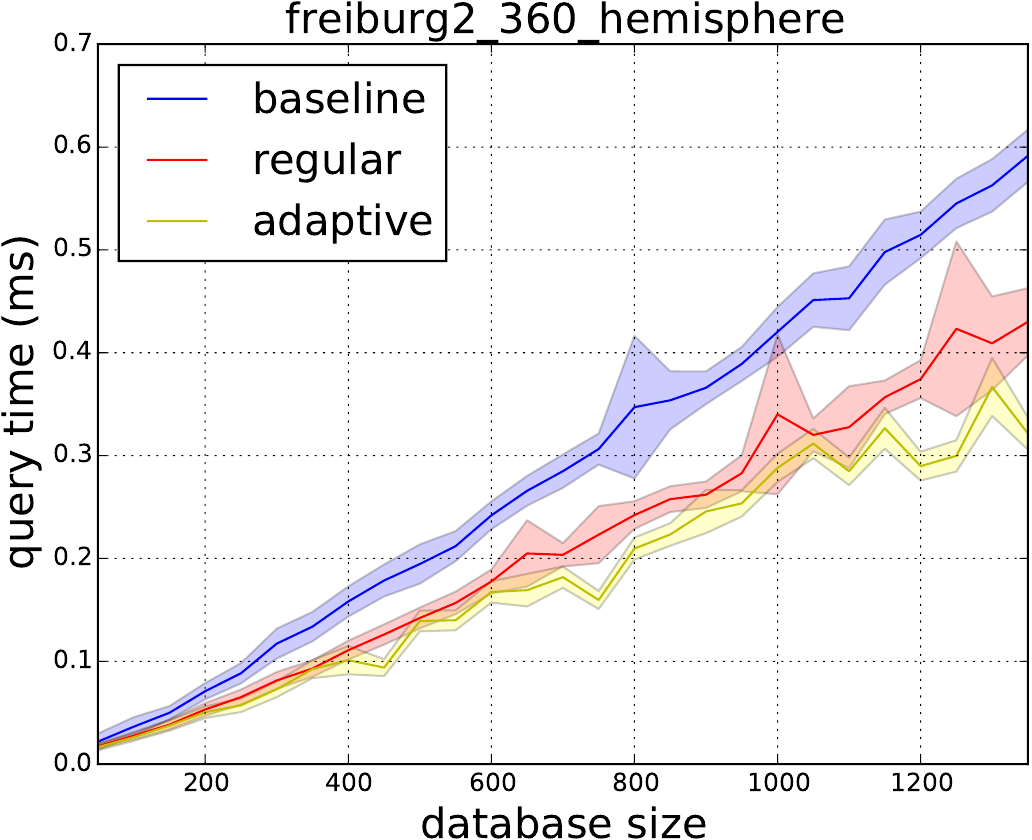} } 
    }
    \subfloat
    {
      {\includegraphics[width=0.3\linewidth]{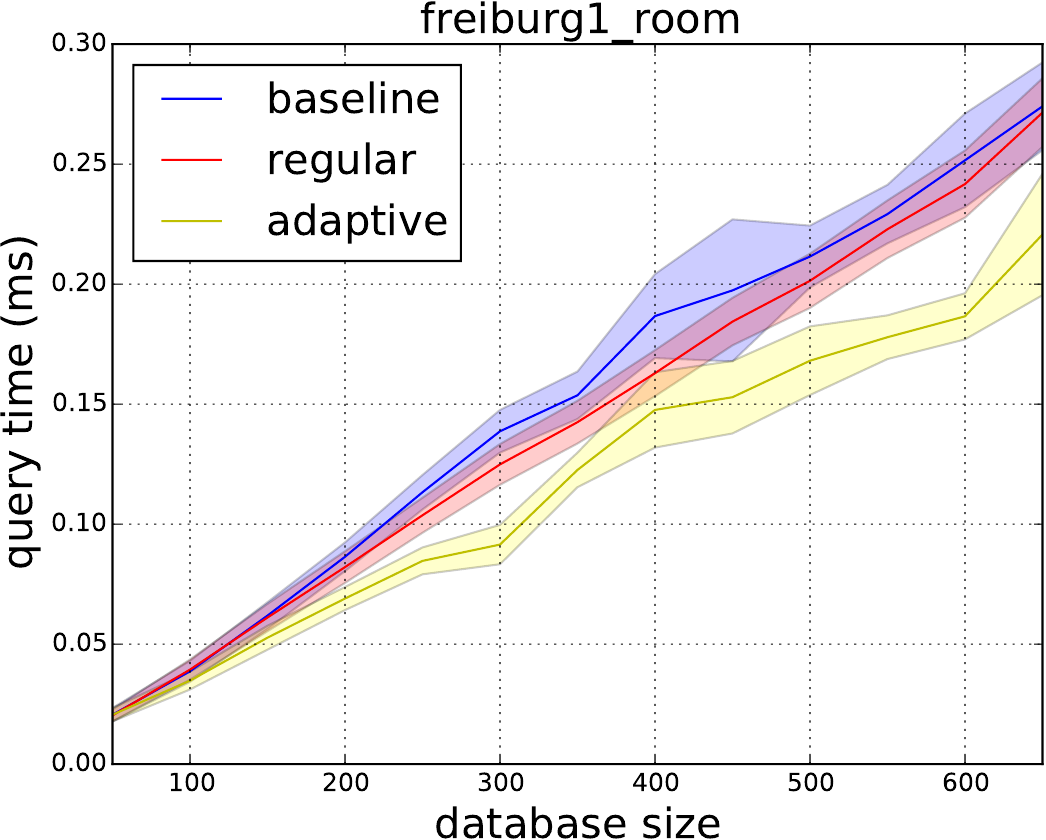} }
    }
    \subfloat
    {
      {\includegraphics[width=0.3\linewidth]{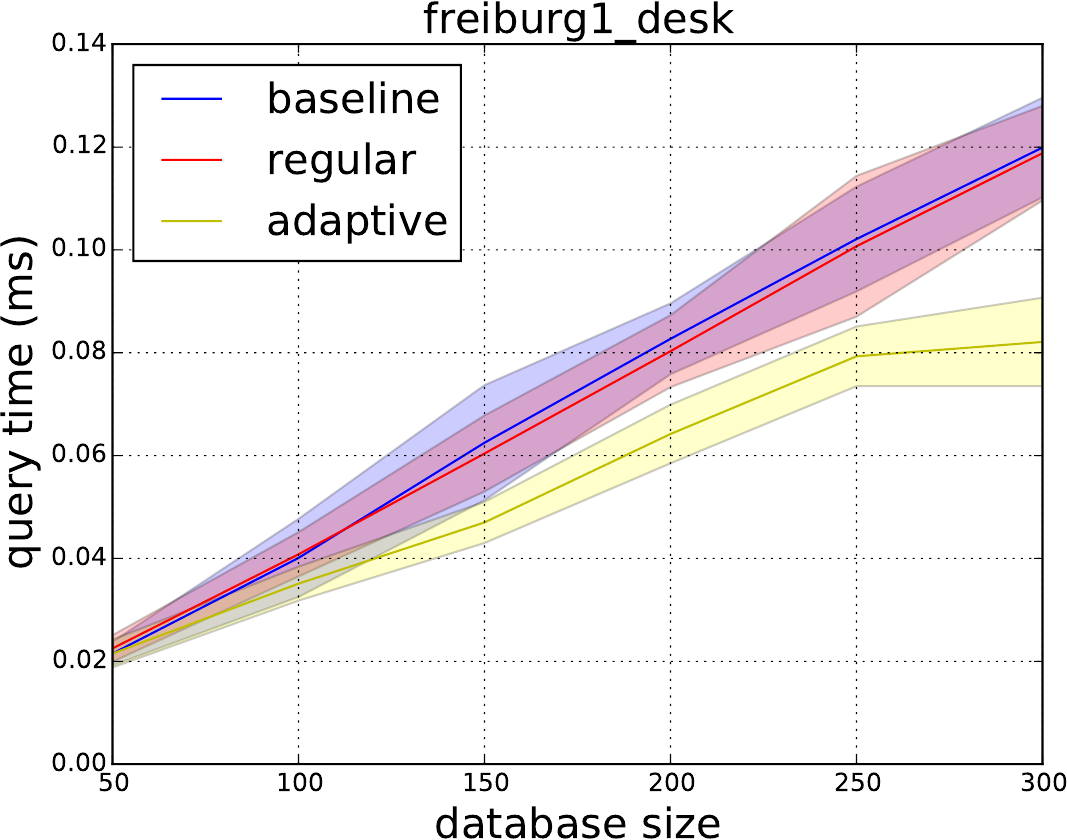} }
    }
  \caption{Sample results on the {\em TUM RGB-D} dataset using adaptive domain clustering (Sec.~\ref{sect-fancy}). The experiment setup is similar to that for the {\it Oxford} dataset in Sec.~\ref{sect-synthex}. Adaptive (yellow) improves with more exciting motion (left to right, up to down). Limited speedup relative to baseline due to very small dataset size. Variance shown is derived from multiple trials with sightly differing cluster assignments.}
\label{fig-TUM}
\end{figure}

Instead of a fixed topology of the data structure, corresponding to regular grouping, we can consider adaptive grouping based on a variety of criteria. Adaptive sampling, or grouping, based on {\em geometry} includes performing spectral clustering in $SE(3)$. Curves in Fig.~\ref{fig-timings} indicate that adaptive sampling achieves marginal improvements compared to regular sampling at a constant rate equal to the average of the adaptive sampling rate. Similarly, parallax-based sampling, based on clustering only the translational component of pose, also yields underwhelming improvements. We do, however, expect adaptive sampling to win in some cases, as it has in a number of smaller-scale experiments we conducted with different motion characteristics from smooth driving, for instance the {\it TUM RGB-D} dataset (Fig.~\ref{fig-TUM})~\cite{sturm12iros}. 


\cut{\subsubsection{Adaptive range-based clustering}
\label{sect-range}
Instead of clustering descriptors based on the domain where they are defined in a loop-closure system, we can cluster descriptors based on where they stand in descriptor space. This is typically done in a discriminative setting using Agglomerative Information Bottleneck (AIB)~\cite{vlfeat,tishbyPB00}, which can also be used to reduce the complexity of the dictionary we use to compose the BoW~\cite{fulkersonVS08}. Unfortunately, the scale of operations desired prevents straightforward application of agglomerative clustering such as AIB, as well as standard data structures such as the KD-Tree or decision trees.
Also, dimensionality reduction by clustering in descriptor space is counter-productive, because the goal is to discriminate pose, and therefore similar descriptors that refer to distant points in pose space should not be averaged (think of driving through a countryside where the landscape changes slightly over long stretches of open road), whereas different descriptors generated by nearby locations (think of rotating in place in a cluttered indoor scene) can be safely aggregated because loop closure can be called so long as any of them triggers a match.}
\subsection{Quantifying speedup using synthetic ground-truth}
\label{sect-synthex}
Depending on the particular query image, our method could reduce or increase search time relative to the mean. The former occurs when correspondence fails early allowing us to rule out subsequent tests at finer scales. However, in the worst-case we may end up performing more comparisons than inverted index search when the test reaches the finest scale too often. In practice, what matters is that our algorithm shortens test time {\em on average} during long sequences. Since most {\em KITTI} sequences contain few or no loop closures, we generate synthetic positive and negative queries as follows: For sequences 01 to 21, we generate positive queries by sampling the right stereo images of each sequence (slightly different from the left images from which we constructed the database), and generate negative queries by sampling images from sequence 00. For the {\em Oxford} datasets, we construct the database using odd-numbered images, generate positive queries from the even-numbered images, and negative queries again from {\em KITTI} 00.

Overall performance is measured by combining both sets of queries. Of course, even in the negative case our algorithm could find erroneous correspondences, which are then labeled as false alarms. Similarly, we may find no correspondence in the former case (missed detection). We use average time-cost rate to evaluate how the searching algorithm scales with size of the database. Tab.~\ref{tab-kittiRegular} reports experiment results on raw {\em KITTI}. Tab.~\ref{tab-keyframe} reports average speedup when keyframe selection is applied on both {\em KITTI} and {\em Oxford}.
\setlength{\tabcolsep}{2pt}
\begin{table}[htb]
 \centering
 \caption{Average time-cost rate and speedup over 21 sequences of {\em KITTI} using \emph{all} frames. 1st col: grouping strategies. 2nd col: pooling operations. 3rd col: average time-cost rate, which describes how the query time increases per 1k images inserted into the database. In \ref{tab-kittiPositive}, \ref{tab-kittiNegative} and \ref{tab-kittiOverall}, a 10K vocabulary is used; in \ref{tab-kittiOverall-1M}, a 1M vocabulary is used.}
 \begin{scriptsize}
 \subfloat[{\small positive queries; KITTI - 10K}]{
 \begin{tabular}{lccc}
 \toprule
structure       & pooling    & rate(ms/1k) & speedup \\
\midrule
inverted index  & N/A        & 10.07  & 1.00 \\
\hline
\multirow{3}{*}{hierarchical}   		& mean       & \textbf{0.69}  &  \textbf{14.59} \\
  		& sum        & 8.70  &  1.16 \\
           		& max        & 6.65  &  1.52 \\
\bottomrule
\end{tabular}
\label{tab-kittiPositive}
  }
  \hspace{\tabspacing}
 \subfloat[{\small negative queries; KITTI - 10K}]{%
 \begin{tabular}{lccc}
\toprule
structure       & pooling    & rate(ms/1k) & speedup \\
\midrule
inverted index  & N/A        & 9.86 & 1.00 \\
\hline
\multirow{ 3}{*}{hierarchical}   		& mean       & \textbf{0.34}  &  \textbf{29.00} \\
  		& sum        & 6.28  &  1.57 \\
           		& max        & 5.04  &  1.96 \\
\bottomrule
\end{tabular}
\label{tab-kittiNegative}
  }\\
 \subfloat[{\small overall; KITTI - 10K}]{%
 \begin{tabular}{lccc}
\toprule
structure       & pooling    & rate(ms/1k) & speedup \\
\midrule
inverted index  & N/A        & 9.88  & 1.00 \\
\hline
\multirow{3}{*}{hierarchical}   		& mean       & \textbf{0.38}  & \textbf{26.00} \\
  		& sum        & 7.92  & 1.25 \\
           		& max        & 6.06  & 1.63 \\
\bottomrule
\end{tabular}
\label{tab-kittiOverall}
  }
  \hspace{\tabspacing}
  \subfloat[{\small overall; KITTI - 1M}]{
 \begin{tabular}{lccc}
 \toprule
structure       & pooling    & rate(ms/1k) & speedup \\
\midrule
inverted index  & N/A        &   0.64 & 1.00 \\
\hline
\multirow{3}{*}{hierarchical}   		& mean       & N/A  & N/A \\
  		& sum        &   \textbf{0.30} &  \textbf{2.13} \\
           		& max        &   \textbf{0.30} &  \textbf{2.13} \\
\bottomrule
\end{tabular}
\label{tab-kittiOverall-1M}
}
\end{scriptsize}
\label{tab-kittiRegular}
\end{table}
Fig.~\ref{fig-kittiCat} shows linear scaling of average query time on the much larger {\em concatenated KITTI}. Practical deployment on even larger datasets typically comes with context ({\it e.g.} GPS or odometry) that limits the data volume. 
\begin{table}[hbt]
\centering
\caption{A comparison of search in flat and hierarchical structure on {\em KITTI} and {\em Oxford} dataset. Notations have the same meanings as in Tab.~\ref{tab-kittiRegular} except that 3rd column describes average time-cost rate over the 21 {\em KITTI} {\em keyframe} sequences and all 4 sequences in the {\em Oxford} dataset respectively. The keyframes are generated by running ORB-SLAM. }
\begin{scriptsize}
 \subfloat[{\small overall; KITTI - 10K}]{
 \begin{tabular}{lccc}
 \toprule
structure       & pooling    & rate(ms/1k) & speedup \\
\midrule
inverted index  & N/A        & 8.97   & 1.00 \\
\hline
\multirow{3}{*}{hierarchical}   		& mean       & \textbf{0.88} &  \textbf{10.14} \\
  		& sum        & 7.87 &  1.14 \\
           		& max        & 6.00 &  1.50\\
\bottomrule
\end{tabular}
\label{tab-kittiKeyframe}
  }
  \hspace{\tabspacing}
 \subfloat[{\small overall; Oxford - 10K}]{%
 \begin{tabular}{lccc}
 \toprule
structure       & pooling    & rate(ms/1k) & speedup \\
\midrule
inverted index  & N/A        & 6.98  & 1.00 \\
\hline
\multirow{3}{*}{hierarchical}   		& mean       & \textbf{1.61}  &  \textbf{4.34} \\
  		& sum        & 4.71  &  1.48 \\
           		& max        & 4.20  &  1.66 \\
\bottomrule
\end{tabular}
\label{tab-OxfordRegular}
  }
\label{tab-keyframe}
\end{scriptsize}
\end{table}
\cut{
\subsection{Empirical Assessment}
\label{sect-assessment}
In general, it seems that the domain-specific constraints would favor deep structures with small branching factors for the purpose of accelerating the query. Assuming that loop closures are rare, which is the case in long driving sequences, we can reject queries by just visiting a few top-level nodes of the hierarchy which is cheaper compared to visiting a lot of bottom-level nodes. However, with an inverted index, querying a node consists only of retrieving sparse common words shared by the query and database entries, which makes visiting extra nodes inexpensive. On the other hand visiting an extra layer of the hierarchy always imposes overhead, which might not be negligible depending on the dataset. Therefore, when an inverted index is employed, a shallow structure with suitable branching factor depending on the application is preferred. Furthermore, for short-term loop closure, as employed during tracking and mapping, the priorities are different. But since in that case scale is not as cogent an issue, we focus on large-scale, rare-loop scenarios. 
Depending on the application context and constraints, one may favor one architecture over another. This determination has to be made empirically. 
}
\subsection{Experiments in image retrieval tasks}
\label{sect-retrieval}
Although our approach is geared towards the loop closure scenario, its usage is not restricted to it. A hierarchical structure of this form could be built on top of any histogram-based representation of images where some proxy of topology is available. In more general settings when a temporal stream of images is unavailable, extra labeling information, such as geotags, class labels, or textual annotations could be used. A hierarchy can be constructed using affinity between these alternate forms of metadata, provided that affinity implies proximity in the solution space. We test this using two publicly available image retrieval benchmarks: {\em ukbench}~\cite{nister2006scalable} and {\em INRIA Holidays}~\cite{jegou2008hamming}.

\noindent{\bf ukbench}~\footnote{\scriptsize\url{http://vis.uky.edu/~stewe/ukbench/}} consists of 2550 groups of 4 images each (10200 total). Each group contains the same object under different viewpoint, rotation, scale and lighting conditions. We use the same evaluation protocol provided by the author: Count how many of 4 images are top-4 when using a query image from that set of four images. We use pre-computed visual words provided by the authors, which are quantized SIFT descriptors using a 1M vocabulary. 

\noindent{\bf INRIA Holidays}~\footnote{\scriptsize\url{https://lear.inrialpes.fr/~jegou/data.php}} contains 500 image groups (1491 total), each of which represents a distinct scene under different rotations, viewpoint and illumination changes, blurring, etc. Performance is measured by mean average precision (mAP) averaged over all 500 queries. We use the 4.5 million SIFT descriptors and 100K vocabulary provided by the authors.

\setlength{\tabcolsep}{1.4pt}
\begin{table}[htb]
\centering
\caption{A comparison of search in flat and hierarchical structure on {\em ukbench} and {\em INRIA Holidays}. 1st col: grouping strategies. 2nd col: pooling operations. 3rd col: average query time. {\em ukbench} takes average number of top-4 retrieved images as score. {\em INRIA Holidays} takes mAP as evaluation metric.}
 \begin{scriptsize}
 \subfloat[{\small ukbench}]{
 \begin{tabular}{lcccc}
\toprule
structure     & pooling & time(ms) & speedup & score \\
\midrule
inverted index   & N/A        & 1.47 &  1.00 & 2.72  \\
\hline
\multirow{3}{*}{\begin{minipage}{40pt}Random\\hierarchical\end{minipage}}     & mean       & 0.38 &  3.87 & 2.80  \\
   & sum        & \textbf{0.37} &  \textbf{3.97} & \textbf{2.83}  \\
           & max        & 0.39 &  3.77 & 2.82  \\
\hline
\multirow{3}{*}{\begin{minipage}{40pt}Greedy\\affinity\\hierarchical\end{minipage}}     &mean       & 0.38 &  3.87 & 2.80  \\
   &sum        & 0.38 &  3.87 & \textbf{2.83}  \\
   &max        & \textbf{0.37} &  \textbf{3.97} & 2.82  \\
\bottomrule
\end{tabular}
\label{tab-ukbench}
  }
  \hspace{\tabspacing}
 \subfloat[{\small INRIA Holidays}]{%
 \begin{tabular}{lcccc}
\toprule
structure     & pooling & time(ms) & speedup & mAP \\
\midrule
inverted index   & N/A        & 9.11 &  1.00 & 0.56  \\
\hline
\multirow{3}{*}{\begin{minipage}{40pt}Random\\hierarchical\end{minipage}}     & mean       & \textbf{5.57} &  \textbf{1.63} & 0.58  \\
   & sum        & 6.19 &  1.47 & \textbf{0.63} \\
           & max        & 6.24 &  1.46 & 0.62  \\
\hline
\multirow{3}{*}{\begin{minipage}{40pt}Greedy\\affinity\\hierarchical\end{minipage}}     &mean       & \textbf{5.58} &  \textbf{1.63} & 0.57  \\
   &sum        & 6.82 &  1.34 & \textbf{0.63}  \\
 &max        & 6.53 &  1.40 & 0.62  \\
\bottomrule
\end{tabular}
\label{tab-holiday}
  }
\label{tab-retrieval}
\end{scriptsize}
\end{table}
The baseline remains to search using an inverted index system. We use a three-layer hierarchy with the original histograms at the bottom layer. At the second layer, histograms belonging to the same object/scene are pooled (pooling based on prior information available about the data and problem space). At the top layer, we compare two different strategies to build the hierarchy: Random grouping and greedy affinity grouping. Random grouping: We randomly group every $N$ histograms from the second layer.  Greedy affinity grouping: We greedily group every $N$ histograms based on their nearest neighbors in affinity (which is the histogram intersection score). In each setup, we also compare the different choices of pooling operators. Tab.~\ref{tab-ukbench} and Tab.~\ref{tab-holiday} show results on the {\em ukbench} and {\em INRIA Holidays} datasets with $N=16$.

In these image retrieval tasks, we {\it completely discard the threshold and only search down those nodes which have top 10 highest scores}. Thus even for sum/max-pooling, the precision-recall behavior should be different from the baseline. All hierarchical approaches, regardless of pooling operation and grouping scheme, are faster than the baseline. The observation that speedup is available even for the random grouping scheme shows that the speedup does not just hinge on grouping similar images, though grouping similar images can boost the speedup further as we have shown in previous experiments on the driving data. We also notice improved score/mAP in these two experiments, likely due to the grouping of histograms of the same object/scene at the second layer of our hierarchy and the top-4 scoring mechanism imposed by the benchmark.
\section{Discussion}
We have presented a hierarchical data structure consisting of pooled local descriptors representing the likelihood of locations given the images they generate, while maintaining an inverted index at each level of the data structure. While mean-pooling of histograms may seem counter-productive, it is a sensible choice when considered an anti-aliasing procedure in the context of classical sampling theory, where the data structure, as well as keyframes, are tasked with {\em down-sampling} the native rate. We have compared several pooling strategies, and found that mean-pooling provides the most speedup at a small cost to performance; sum-pooling has the upper-bound property and accelerates search to a reasonable degree without loss of performance; and max-pooling shares the same property with sum-pooling but exhibits a larger speedup due better approximating the nodes below it.

For simplicity, we chose a fixed topology (depth and branching factor) and studied the resulting performance empirically. We have found that sophisticated heuristics do not improve performance enough to justify the added complexity. We have benchmarked our scheme on public datasets, where we have shown that even a shallow tree can significantly cut down on test time with minimal impact to precision, which is the main goal of loop closure. 
\section*{Acknowledgements}
This work was supported by AFRL FA8650-11-1-7156, ONR N00014-15-1-2261 and ARO W911NF-15-1-0564.


\clearpage

\bibliographystyle{splncs03}

\end{document}